\documentclass[runningheads]{llncs}

\usepackage{eccv}

\usepackage{eccvabbrv}
\usepackage[ruled]{algorithm2e}
\usepackage{graphicx}
\usepackage{booktabs}

\usepackage{amsmath}
\usepackage{multirow}
\usepackage{makecell}
\usepackage{pifont}

\usepackage[accsupp]{axessibility}  

\usepackage{hyperref}

\usepackage{orcidlink}

\begin{document}

\title{SpeedUpNet: A Plug-and-Play Adapter Network for Accelerating Text-to-Image Diffusion Models} 

\titlerunning{SpeedUpNet}

\author{Weilong Chai\orcidlink{0009-0003-0038-653X}\thanks{Equal contribution.} \and
Dandan Zheng\orcidlink{0009-0005-2151-3547}$^{\star}$ \and
Jiajiong Cao\orcidlink{0000-0001-8311-5820} \and
Zhiquan Chen\orcidlink{0000-0001-8382-9524} \and\\
Changbao Wang\orcidlink{0009-0009-9870-3612} \and
Chenguang Ma\orcidlink{0000-0002-3627-2740}
}

\authorrunning{Chai et al.}

\institute{Ant Group \\
\email{\{weilong.cwl,yuandan.zdd,jiajiong.caojiajio,\\zhiquan.zhiquanche,changbao.wcb,chenguang.mcg\}@antgroup.com}}

\maketitle

\begin{abstract}

Text-to-image diffusion models (SD) exhibit significant advancements while requiring extensive computational resources. 
Existing acceleration methods usually require extensive training and are not universally applicable. 
LCM-LoRA, trainable once for diverse models, offers universality but rarely considers ensuring the consistency of generated content before and after acceleration. 
This paper proposes SpeedUpNet (SUN), an innovative acceleration module, to address the challenges of universality and consistency. 
Exploiting the role of cross-attention layers in U-Net for SD models, we introduce an adapter specifically designed for these layers, quantifying the offset in image generation caused by negative prompts relative to positive prompts. 
This learned offset demonstrates stability across a range of models, enhancing SUN's universality. 
To improve output consistency, we propose a Multi-Step Consistency (MSC) loss, which stabilizes the offset and ensures fidelity in accelerated content. 
Experiments on SD v1.5 show that SUN leads to an overall speedup of more than \textbf{10} times compared to the baseline 25-step DPM-solver++, and offers two extra advantages: (1) training-free integration into various fine-tuned Stable-Diffusion models and (2) state-of-the-art FIDs of the generated data set before and after acceleration guided by random combinations of positive and negative prompts.
Code is available\footnote[1]{Project: https://williechai.github.io/speedup-plugin-for-stable-diffusions.github.io}.
\keywords{Diffusion models \and Acceleration \and Adapter network}
\end{abstract}

\section{Introduction}
In recent years, significant advancements have been made in the field of generative models, particularly in text-to-image generation, with Denoising Diffusion Probabilistic Models (DDPMs) \cite{ho2020denoising} playing a crucial role. 
To further enhance the generation quality of text-to-image diffusion models, classifier-free guidance (CFG) \cite{ho2022classifier} is widely used in large-scale generative frameworks \cite{nichol2021glide} \cite{rombach2022high} \cite{ramesh2021zero} \cite{saharia2022photorealistic}. 
However, the iterative sampling procedure for diffusion models costs extensive computational resource, and CFG doubles the inference latency because it demands one diffusion process for the positive prompt and another for the negative.

Based on the above problems, many efforts have been made on the topic of fast sampling and distillation of diffusion models. 
Advanced sampling strategies \cite{song2020denoising} \cite{lu2022dpm} \cite{lu2022dpm++} significantly decrease the diffusion steps from several hundreds to $25$ without training. 
Structural pruning \cite{kim2023architectural} proposes that that a smaller ``student" model can be trained to mimic the output of the ``teacher" model.
To reduce the inference steps of the diffusion model, Progressive distillation \cite{salimans2022progressive} and Consistency Models \cite{song2023consistency} learn to iteratively reduce the sampling steps.
Guided-Distill \cite{meng2023distillation} and Latent Consistency Models (LCM) \cite{luo2023latent} \cite{luo2023lcm} augment above methods to text-to-image diffusion models, where CFG process is particularly considered in their distillation processes. 

These methods have shown producing high-quality images in less than 4 sampling steps, but they still have several limitations in practicality. First, existing distillation methods require fine-tuning of the entire diffusion network and corresponding training data, which makes them difficult to apply to a new pre-trained model. 
Second, efficient finetuning methods that use LoRA \cite{luo2023lcm} may result in significant visual differences 
between the images generated before and after acceleration.
This is also accompanied by the inaccuracy of the indicators, when the selected dataset (LAION5B or MCOCO) for FID and CLIP-score can be very different in terms of data distribution from the dataset used for training the stylized SD model.
Additionally, input from negative prompts is often simplified or discarded during accelerations, which weakens the adjustability of the accelerated models.

\begin{figure}[t] \centering
    \includegraphics[width = 0.7\textwidth]{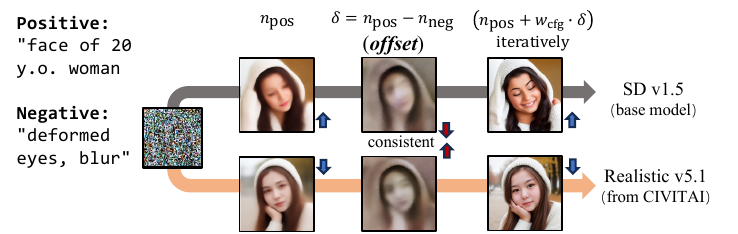}
    \caption{
      Visualization of offset between positive and negative guidances.
      While finetuned SD can generate images of very different styles, the substraction of predictions guided by positive and negative text (\textit{offset}) is relatively \textbf{consistent} in different SDs.
    }
    \label{fig:image_vis}
  \end{figure}

To address these limitations, we propose a novel and universal acceleration adapter called SpeedUpNet (SUN). 
Once trained on a base Stable Diffusion (SD) \cite{rombach2022high} model, SUN can be easily plugged into various fine-tuned SD models (such as different stylized models) to significantly improve inference efficiency while maintaining content consistency and negative prompt control.
In particular, SUN is implemented in a teacher-student distillation framework, where the student has the same architecture with the teacher model except for an additional adapter network. 
During the training, only the adapter of the student, which consists of several cross-attention layers, are optimized with the other parameters frozen. 
The adapter network takes the negative prompt embedding as an extra input to the diffusion model, allowing for CFG-like effects in one inference. 
SUN adapter consists of several cross-attention operations to calculate the offset of the negative prompt relative to the positive prompt on each of the attention layers in the U-Net. 
As depicted in Fig.\ref{fig:image_vis}, the offset $\delta$ between negative and positive text embeddings, which is usually utilized to improve image quality,
is noted to be a variable associated with text inputs and is unrelated to the model's style.
As a result, the trained adapter network can be generalized to other stylized T2I diffusion models.


Additionally, SUN introduces a Multi-Step Consistency (MSC) loss to ensure a harmonious balance between reducing inference steps and maintaining consistency in the generated output. 
Different from the existing method that gradually change the inverse diffusion trajectory to a new one for one-step generation such as LCM \cite{luo2023latent} and Guided-Distill \cite{meng2023distillation}, 
MSC divides the original dense trajectory into a few (e.g. 4) stages, with each stage being approached by an accelerated inference. By mapping the output of each stage to the point of the original trajectory, this method avoids cumulative errors during acceleration, thus maintaining the consistency of the output image.
Consequently, SUN significant reduces in the number of inference steps to just 4 steps and eliminates the need for CFG, which leads to an overall speedup of more than 10 times for SD models compared to the 25-step dpm-solver++.
To sum up, our contributions are as follows:
\begin{itemize}
  \item First, we propose a novel and universal acceleration module called SpeedUpNet (SUN), which can be seamlessly integrated into different fine-tuned SD models without training, once it is trained on a base SD model. 
  \item Second, we propose a method that supports classifier-free guidance distillation with controllable negative prompts and utilizes Multi-Step Consistency (MSC) loss to enhance content consistency between the generated outputs before and after acceleration.
  \item Third, experimental results demonstrate SUN  achieves a remarkable speedup of over 10 times on diffusion models. SUN fits various style models as well as generation tasks (including Inpainting \cite{meng2021sdedit}, Image-to-Image and ControlNet \cite{zhang2023adding}) without extra training, and achieve better results compared to existing SOTA methods.
\end{itemize}

\section{Related Work}
\subsection{Diffusion Models and Classifier-free Guidance}
Diffusion Models have achieved great success in image generation (\cite{ho2020denoising}\cite{song2020denoising}\cite{nichol2021improved}\cite{ramesh2022hierarchical}). 
Classifier-free guidance (CFG\cite{ho2022classifier}) is an technique for improving the sample quality of text-to-image diffusion models,
which has been applied in models such as GLIDE \cite{nichol2021glide}, Stable Diffusion \cite{rombach2022high} and DALL·E 2 \cite{ramesh2021zero}. 
It incorporates a guidance weight that balances the trade-off between sample quality and diversity during the generation process. 
However, it should be noted that this approach increases the computational load due to the requirement of evaluating both conditional and unconditional (positive and negative prompts) models at each sampling step, thus necessitating optimization strategies to improve speed.

\subsection{Accelerating Diffusion Models}

The advanced sampling strategies (including DDIM \cite{song2020denoising}, DPM-Solver \cite{lu2022dpm} and DPM-Solver++ \cite{lu2022dpm++}) significantly decrease the number of diffusion steps from several hundreds to around 25. 
On structural pruning,
BK-SDM \cite{kim2023architectural} introduces different types of efficient diffusion models and propose various distillation strategies. 
On step distillation,
Progressive Distillation (PD) \cite{salimans2022progressive} and Guided-Distill \cite{meng2023distillation} propose progressive distillation methods, where a student model can generate high-quality images with only 2 diffusion steps. 
Additionally, Consistency Models (CM) \cite{song2023consistency} generate image in a single step by utilizing consistency mapping derived from ODE trajectories. 
Based on CM, Latent Consistency Models (LCM) \cite{luo2023latent} is proposed for accelerating text-to-image synthesis tasks. 
Some recent studies, such as UFOGen \cite{xu2023ufogen} and ADD \cite{sauer2023adversarial} use adversarial techniques to obatain high-quality images in fewer steps. 
By incorporating LoRA into the distillation process of LCM, without fine-tuning the entire network, LCM-LoRA \cite{luo2023lcm} achieves a reduction in the memory overhead of distillation, 
as well as the ability for accelerating diverse models and tasks.

\begin{figure} [t]
    \centering
    \includegraphics[width=11cm]{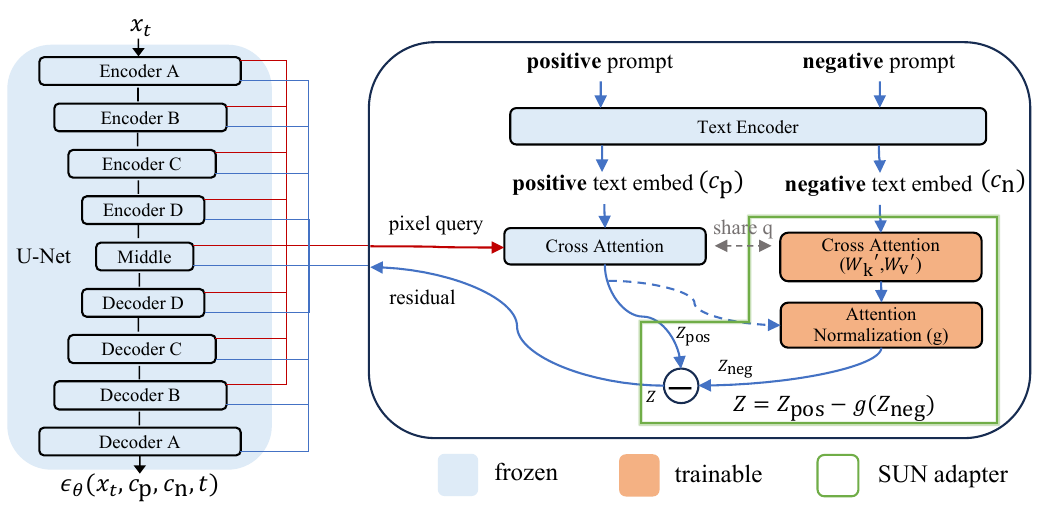}
    \caption{The overall framework of the proposed SUN. 
    SUN adapter is introduced to process and understand the negative prompt, which consists of several cross attention (CA) blocks.
    Each CA of SUN is placed side by side on each block of the original U-Net. 
    Each block introduces a new K matrix and a V matrix, while sharing the Q with the original U-Net. 
    Attention Normalization technique is proposed for stablability.
    }
    \label{fig:framework}
  \end{figure}

\subsection{HyperNetwork and Adapter Methods}
These approaches primarily focus on fine-tuning pre-existing models for specific tasks without extensive retraining. 
HyperNetworks \cite{Hypernetwork2022style}, with the aim of training a small recurrent neural network to influence the weights of a larger one, 
have found their way into adjusting the behavior of GANs and diffusion models. 
To retrofit existing models with new capabilities, adapters have been shown effective in vision-language tasks and text-to-image generation.
ControlNet \cite{zhang2023adding} tailors SD output by conditioning. T2I-adapters \cite{mou2023t2i} offers fine-tuned control over attributes such as color and style. 
IP-Adapter \cite{ye2023ip}, which is an efficient and lightweight adapter, enables image prompt capability for pretrained text-to-image diffusion models.  

\section{Method} \label{sec:optimization}
\subsection{Problem Formulation}
The latent diffusion process can be inferred by optimizing the subsequent equation:
\begin{equation}
\mathcal{L} = \mathbb{E}_{\boldsymbol{x},\boldsymbol{\epsilon},\boldsymbol{p},t}[||\epsilon_{\theta}(\boldsymbol{x}_{t},\operatorname{E}(\boldsymbol{p}),t)-\boldsymbol{\epsilon}||_{2}^{2}],
\end{equation}
where $\boldsymbol{x}$ symbolizes the noisy latent representation of an image, $\boldsymbol{p}$ is the corresponding prompt, $\operatorname{E}$ represents the text encoder transforming $\boldsymbol{p}$ to a conditional embedding, and $t$ symbolizes a time step, sampled from a uniform distribution $t\sim \text{Uniform}(0,1)$. The noise $\boldsymbol{\epsilon}$ adheres to a standard Gaussian distribution, i.e., $\epsilon\sim N(0, I)$. During the inference process, two texts, a positive prompt $\boldsymbol{p}_{\text{p}}$ and a negative prompt $\boldsymbol{p}_{\text{n}}$, are applied as conditions of two independent diffusion steps:
\begin{equation}
\begin{aligned}
&\boldsymbol{\epsilon}_{p} = \boldsymbol{\epsilon}_{\theta}(\boldsymbol{x}_{t},\operatorname{E}(\boldsymbol{p}_{\text{p}}),t), \\
&\boldsymbol{\epsilon}_{\text{n}} = \boldsymbol{\epsilon}_{\theta}(\boldsymbol{x}_{t},\operatorname{E}(\boldsymbol{p}_{\text{n}}),t), \\
&\hat{\boldsymbol{\epsilon}} = w \boldsymbol{\epsilon}_{\text{p}} + (1-w)\boldsymbol{\epsilon}_{\text{n}},
\end{aligned}
\end{equation}
where $\boldsymbol{\epsilon}_{\text{p}}$, $\boldsymbol{\epsilon}_{\text{n}}$, and $\hat{\boldsymbol{\epsilon}}$ represent the positive noise, negative noise, and the final noise, respectively. This process requires two forwards of the model in order to compute the final noise, leading to potential computational inefficiency.
In this study, we propose a strategy to predict the final noise in a single forward pass. 
\begin{equation}
\begin{aligned}
\hat{\boldsymbol{\epsilon}} = \boldsymbol{\epsilon}_{\theta}(\boldsymbol{x}_{t},\operatorname{E}(\boldsymbol{p}_{\text{p}}),\phi(\operatorname{E}(\boldsymbol{p}_{\text{n}})),t),
\end{aligned}
\end{equation}
where $\phi$ is a decoupled network with $\boldsymbol{\epsilon}_{\theta}$ and $\operatorname{E}$, which can be optimized independently.

\subsection{Negative-positive Offset Learning}
The overall framework and details of our method is illustrated in Figure \ref{fig:framework}. 
During training, the negative text embedding is fed into the SUN adapter, which consists of several cross-attention blocks. 
And the SUN adapter interacts with the original U-Net through the cross-attention blocks. 
During inference, the SUN adapter can be directly plugged into any fine-tuned SDs. 

To embody the interaction amongst $\epsilon_{\theta}$, $E$ and the proposed decoupled network $\phi$, a cross-attention mechanism is integrated. The original interaction between $\epsilon_{\theta}$ and $E$ can be expressed as follows:
\begin{equation}
\begin{aligned}
Z_{\text{p}} = \operatorname{MHSA}(Q, K, V),
\end{aligned}
\label{eq:raw_mhsa}
\end{equation}
where $\operatorname{MHSA}$ refers to the multi-head self-attention operation, $Q=Z W_{\text{q}}$, $K=E(p_{\text{p}}) W_{\text{k}}$, $V=E(p_{\text{p}}) W_{\text{v}}$ are the query, key, and value matrices of the attention operation. Respectively, $W_{\text{q}}$, $W_{\text{k}}$, $W_{\text{v}}$ represent the weight matrices of the trainable linear projection layers.
To insert negative text embedding, a new attention operation is added for the decoupled network $\phi$:
\begin{equation}
\begin{aligned}
Z_{\text{n}} = \operatorname{MHSA}(Q, K', V'),
\end{aligned}
\end{equation}
where $Q$ is shared from Equation \ref{eq:raw_mhsa}, 
and $K'=\operatorname{E}(p_{\text{n}}) W'_{\text{k}}$, $V'=\operatorname{E}(p_{n}) W'_{\text{v}}$ represent the key and value of the negative text embedding.

\begin{figure}[tb] \centering
    \includegraphics[width = 0.75\linewidth]{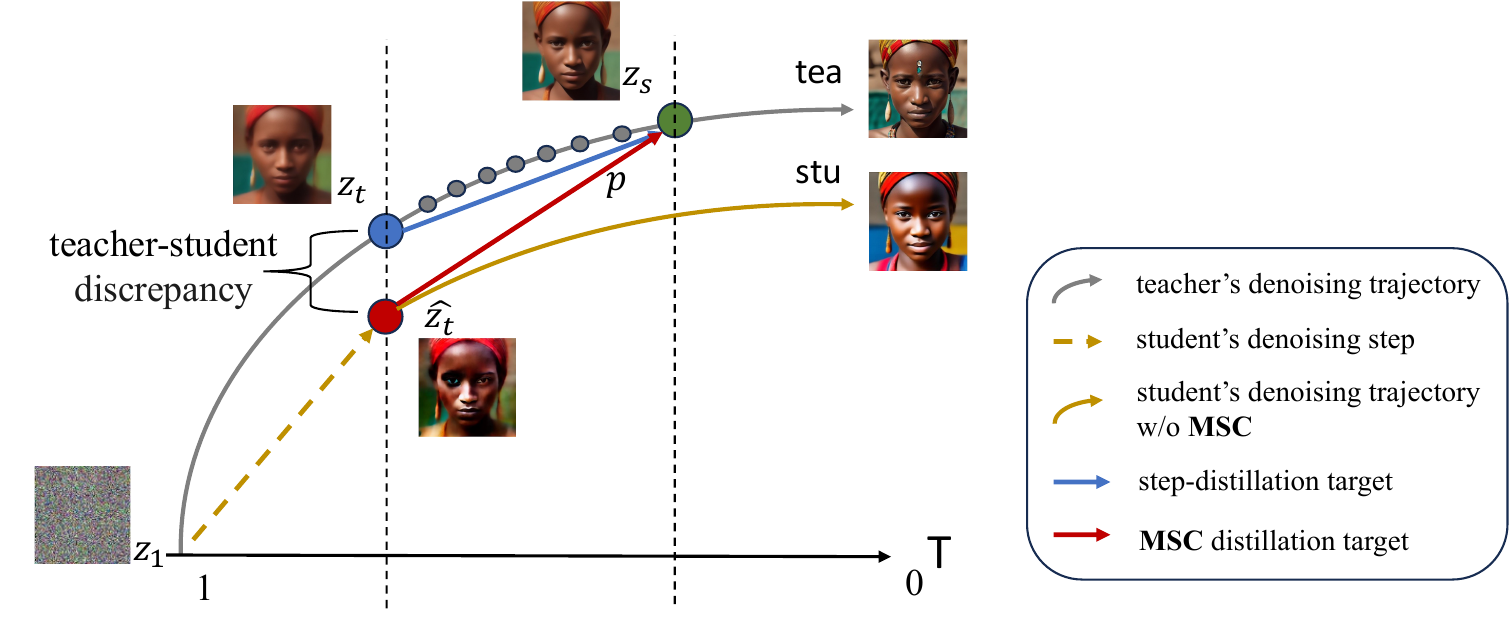}
    \caption{
        An illustraion of Muiti-step Consistency (MSC). When distilling a faster student model, teacher-student discrepancy exists and gradually accumulates, causing the content generated by the student to be inconsistent with the teacher (from the same noise). Based on the step distillation method, MSC is used to train the student to approach the teacher's trajectory even when error occurs, thus ensuring consistency in muiti-step samplings.
    }
    \label{fig:cfgmsc}
\end{figure}

\begin{algorithm}[tb] 
  \caption{Training of SpeedUpNet}
  \label{alg:stoc}
  \small
  \LinesNumbered
  \KwIn{image-caption dataset $\mathcal{X}$, negative prompt dataset $\mathcal{Y}$, stable diffusion base model $\epsilon_{\theta}$, SUN adapter $\phi$ with parameter $\theta_{\phi}$.}
  \While{not converged}{
  Sample $(x,\boldsymbol{p}_{p}) \sim p(X)$, $\boldsymbol{p}_{n} \sim p(Y)$\;
  Forward $\boldsymbol{c}_\text{p} \leftarrow \text{E}(\boldsymbol{p}_{p})$, $\boldsymbol{c}_\text{n} \leftarrow \text{E}(\boldsymbol{p}_{n})$, $o \leftarrow \boldsymbol{\epsilon}_\theta\left(\boldsymbol{x}_t, \boldsymbol{c}_\text{p}, \boldsymbol{c}_\text{n}, t\right)$\;
  Calculate $\mathcal{L}_{\text{cls}}$ via Eq.~\ref{eq:loss_cfg}\;
  Calculate $\tilde{\boldsymbol{\epsilon}_{\text{step}}}$ via Eq.~\ref{eq:psuedo_epsilon}\;
  Calculate $\mathcal{L}_{\text{msc}}$ on $o$ and $\tilde{\boldsymbol{\epsilon}_{\text{step}}}$ via Eq.~\ref{eq:loss_step} and Eq.~\ref{eq:loss_msc}\;
  Calculate $\mathcal{L}$ via Eq.~\ref{eq:loss_all}\;
  Update $\theta_{\phi}$ with gradient $\nabla_{\theta_{\phi}}(\mathcal{L})$\;
  }
\KwOut{$\theta_{\phi}$}
\end{algorithm}

The computation of the final feature, denoted as Z, is critical to the overall interaction between $\epsilon_{\theta}$, $E$, and $\phi$ as it encapsulates the negative impact of the text $p_{\text{n}}$. This is demonstrated through a subtraction operation:
\begin{equation}
\begin{aligned}
Z = Z_{\text{p}} - g(Z_{\text{n}}),
\end{aligned}
\end{equation}
 where $g$ is a function called \textbf{Attention Normalization} which lies in the necessity to balance the contributions from the positive and negative text prompts and to regulate the scale of the feature vectors. It is defined as:
\begin{equation}
\begin{aligned}
g(Z_{\text{n}}) = \alpha Z_{\text{n}} \times \operatorname{norm}(Z_{\text{p}}) / \operatorname{norm}(Z_{\text{n}}) + \beta,
\end{aligned}
\end{equation}
where $\operatorname{norm}$ is a function that computes the magnitude of a vector, providing an objective measure of the contribution from each feature. The parameters $\alpha$ and $\beta$ are learnable weights that allow the model to adaptively control the strength of influence of the negative prompt. Attention Normalization $g$ helps to improve the generalization of SUN, which we describe in the experiments chapter.

\subsection{Multi-Step Consistency (MSC) Distillation}
\subsubsection{Vanilla CFG Distillation}. 
To imitate the behavior of classifier-free diffusion model, one of the objective is to encourage the output of the student to resemble the prediction by classifier-free guidance:
\begin{equation}
    \label{eq:loss_cfg}
    \mathcal{L}_{\text{cfg}}=\mathbb{E}_{\boldsymbol{x}_0, \boldsymbol{c}_\text{p}, \boldsymbol{c}_\text{n}, t}\left\|\hat{\boldsymbol{\epsilon}}-\boldsymbol{\epsilon}_\theta\left(\boldsymbol{x}_t, \boldsymbol{c}_\text{p}, \boldsymbol{c}_\text{n}, t\right)\right\|^2 \text{,}
\end{equation}
where $\boldsymbol{c}_\text{p}=\text{E}(\boldsymbol{p}_{p})$ is the conditional embedding of the positive prompt, $\boldsymbol{c}_\text{n}=\text{E}(\boldsymbol{p}_{n})$ represents the conditional embedding of the negative prompt, and $\hat{\boldsymbol{\epsilon}}=w \boldsymbol{\epsilon}_{p} + (1-w)\boldsymbol{\epsilon}_{n}$ is the final noise given by teacher's CFG.
It is worth to notice that that there are two differences from the original CFG-Distill \cite{meng2023distillation} method.
First, the original SD model $\theta$ is frozen and only the parameters of the adapter network $\phi$ are optimized.
Second, various negative prompts are used in training instead of a fixed empty prompt to ensure that the model is still controlled by negative prompts when producing content.
These changes make the optimization goal closer to the inference procedure, and make the adapter more versatile.

\subsubsection{Multi-step Consistency Loss}. 
In order to further improve the model to sample high-quality and consistent images in fewer steps, we use optimized step-distillation and add MSC loss on this basis to reduce the gap between the student and the teacher.
As the SUN adapter has already accepted negative prompts as input, there is no need for pre-distillation to remove CFG as done in Guided-Distill \cite{meng2023distillation}. To maintain a stable teacher, we also choose not to progressively distill it multiple times like PD \cite{salimans2022progressive}.
Given the noisy input $\boldsymbol{x}_{t}$ at time $t$ and the teacher's sampling process from time $t$ to $s$ by N steps in continuous time space, 
the objective for the student network is to obtain the same diffusion state $\tilde{\boldsymbol{x}_{s}}$ at time $s$ in one step.
To perform the sampling process in the continuous time space, we divide the time $t$ to $s$ into $N$ segments, and get $\Delta = (t-s)/N$ as the time interval for each inference process.
For $t^{\prime}$ in $\left[t, t-\Delta, t-2\Delta, ..., s+\Delta\right]$ and let $t^{\prime\prime}=t^{\prime}-\Delta$,
the ideal noisy sample $\tilde{\boldsymbol{x}_{s}}$ at time $s$ can be obtained by iteratively inferencing the teacher network via classifier-free guidance:
\begin{equation}
    \label{eq:consist_diffuse}
    \boldsymbol{x}_{t^{\prime\prime}}=\alpha_{t^{\prime\prime}}\frac{\boldsymbol{x}_{t^{\prime}}-\sigma_{t^{\prime}}\hat{\boldsymbol{\epsilon}_{\theta^{t^{\prime}}}}\left(\boldsymbol{x}_{t^{\prime}}\right)}{\alpha_{t^{\prime}}} + \sigma_{t^{\prime\prime}} \hat{\boldsymbol{\epsilon}_{\theta ^{\prime}}}\left(\boldsymbol{x}_{t^{\prime}}\right) \text{.}
\end{equation}
In order for the model to generate $\tilde{\boldsymbol{x}_{s}}$ from $\boldsymbol{x}_{t}$ in one step, the network should predict 
$\tilde{\boldsymbol{\epsilon}_{\text{step}}}$ approximately. According to DDIM updating rule, we have
\begin{equation}
    \label{eq:psuedo_epsilon}
    \tilde{\boldsymbol{\epsilon}_{\text{step}}} = \frac{(\tilde{\boldsymbol{x}_{s}} - \frac{\alpha_s\boldsymbol{x}_{t}}{\alpha_t})}{\sigma_s-\frac{\alpha_s\sigma_t}{\alpha_t}} \text{.}
\end{equation}
The corresponding step-distillation loss is calculated as 
\begin{equation}
    \label{eq:loss_step}
    \mathcal{L}_{\text{step}}(\boldsymbol{x}_t)=\mathbb{E}_{\boldsymbol{x}_0, \boldsymbol{c}_\text{p}, \boldsymbol{c}_\text{n}, t}\left\|\tilde{\boldsymbol{\epsilon}_{\text{step}}}-\boldsymbol{\epsilon}_\theta\left(\boldsymbol{x}_t, \boldsymbol{c}_\text{p}, \boldsymbol{c}_\text{n}, t\right)\right\|^2 \text{.}
\end{equation}
It is important to note that there is a discrepancy between the output of the student network and the teacher network, and this discrepancy will accumulate with iterative sampling, leading to inaccurate results.
To address the issue, we introduce the MSC loss to rectify the step-distillation loss (as shown in Fig. \ref{fig:cfgmsc}). When selecting the value of $\boldsymbol{x}_{t}$, we randomly replace it with the student's output from the previous moment $\hat{\boldsymbol{x}_{t}}$ with a probability of $p$. Regardless of whether the input is sampled from the teacher's
sampling process or the student's, the student is forced to generate $\tilde{\boldsymbol{x}_{s}}$ to ensure that the next moment follows the original trajectory without deviation:
\begin{equation}
    \label{eq:loss_msc}
    \mathcal{L}_{\text{msc}} = \mathcal{L}_{\text{step}}(i\boldsymbol{x}_t + (1-i)\hat{\boldsymbol{x}_{t}}) \text{,}
\end{equation}
where
\begin{equation}
    \label{eq:loss_msc_p}
    P(i=0)=p \text{, } P(i=1)=1-p.
\end{equation}
Considering with CFG-distill loss, the overall optimization target is
\begin{equation}
    \label{eq:loss_all}
    \mathcal{L} = \mathcal{L}_{\text{cfg}} + \lambda \mathcal{L}_{\text{msc}} \text{.}
\end{equation}

The entire training process proceeds as shown in Algorithm \ref{alg:stoc}.

\section{Experiments}

\subsection{Details of Implementation}
\subsubsection{Dataset}.
To train the proposed network,
we use LAION-Aesthetics-6+ which is a sub set of 
LAION-5B \cite{schuhmann2022laion5b}
containing 12M text-image pairs with predicted aesthetics scores higher than 6.
Each sample from the original dataset includes one prompt (deemed a positive prompt) and a corresponding image. 
Subsequently, we leveraged two distinct strategies to collect negative prompts: (1) extracting negative prompts from AIGC websites such as PromptHero \cite{prompthero}; 
(2) utilizing large language models to generate a negative counterpart for the positive prompt.
We then split every negative prompt into phrases with comma, resulting in a total of 832 distinct phrases.
During the training, in order to generate a negative prompt, we uniformly sample 0 to 100 phrases from all the phrases, and then join them into a complete prompt (e.g. "watermark, blurry, ugly, bad anatomy, bad hands, error, missing ﬁngers").

\subsubsection{Configuration for Experiments}. 
We use the widely-used Stable Diffusion v1.5 \cite{rombach2022high} for the base model. In our SUN adapter, we incorporate 16 cross-attention modules that is trainable during the distillation, resulting in a total parameter count of 18.5 M. 
Our method is implemented based on the Diffusers library \cite{diffusers} and PyTorch \cite{paszke2019pytorch}. The training is launched on a single machine with 4 A100 GPUs for approximately 5k steps using the batch size of 32. 
Utilizing acceleration libraries allows the model to be trained on a single machine with 8 V100 GPUs for around 20k steps with a batch size of 8. The results derived from both machine configurations are competitive. 
We utilize the AdamW optimizer, maintaining a constant learning rate of 0.0001 and a weight decay of 0.01. The training process involves resizing the image's shortest side to 512, followed by a 512 $\times$ 512 center crop. 
For MSC loss, $\lambda$ is set to 1.0, $\Delta$ is set 0.25, $p$ is set 0.1.

\subsubsection{Baselines and Evaluation}. \label{sec:evaluation}
For training-free methods, we use DDIM \cite{song2020denoising}, DPM-Solver \cite{lu2022dpm}, and DPM-Solver++ \cite{lu2022dpm++} schedulers.
For training-requiring methods, we compare with Guided-Distill \cite{meng2023distillation} and LCM \cite{luo2023latent}. Since there has been no open-sourced training-required method before, we reproduce Guided-Distill following the paper, on our dataset configurations.
Since our method also belongs to adaptating-free methods that require training only once and can used with other pre-trained models, and we compare it to LCM-LoRA \cite{luo2023lcm}.
Following previous works, we test on LAION-Aesthetics-6+ dataset.
We use FID and CLIP scores to evaluate the performances, where we generate 30K images using
10K text prompts of test set with 3 random seeds.

\begin{figure}[tb] \centering
    \includegraphics[width = 0.9\textwidth]{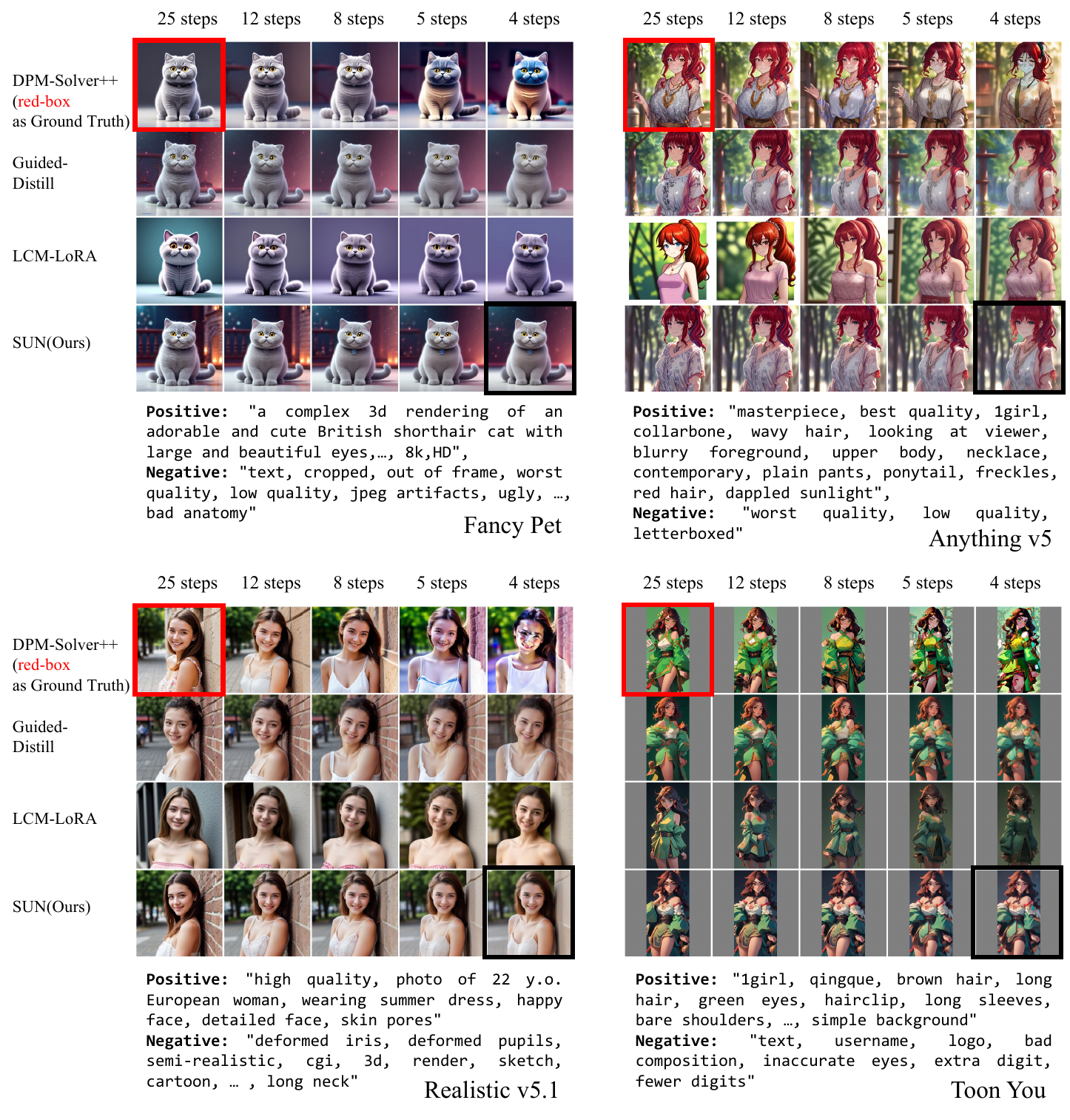}
    \caption{Generation comparisons with different SOTA methods on different numbers of diffusion steps. The proposed SUN can produce high-quality images with only a few steps. In addition, the proposed SUN achieves the highest consistency to the ground truth with only 4 steps.}
    \label{fig:steps}
\end{figure}

\subsection{Qualitative Results}

Without further training, we insert SUN pre-trained based on SD v1.5 into different popular diffusion models from CIVITAI \cite{civitai}, including Anything v5, Realistic Vision v5.1, Toon You, and Fancy Pet. These models all use the same network structure and noise-prediction as SD v1.5.
To display the result, we mainly compare our method with LCM-LoRA, which is also a training-free acceleration method. Additionally, we compare with Guided-Distill, which requires training, to perform further training on each model for comparisons. We regard the results of DPM-solver ++ (25 steps) as ground truths. By reducing the number of inference steps of each method, we compare the difference between its generated results and the ground truth.

As shown in Fig. \ref{fig:steps}, DPM-solver++ produce significantly poorer quality images when using a smaller number of steps (e.g. 4, 8 steps). LCM-LoRA can enhance the quality of images in the above situation without training, but the generated images may vary significantly from the ground truth. In contrast, SUN not only produces high-quality images but also generates consistent results with the ground truth at different choices of sampling steps. This reflects that SUN as a universal acceleration module is more versatile when being plugged into new models.
Compared with the training-hungry method, SUN also has advantages in content consistency, which shows that MSC objective plays a role in reducing student-teacher discrepancies. At the same time, since SUN only trains the adapter parameters, it maintains well the image style and quality from the original model.

\begin{table}[tb] \centering
    \caption{
        Quantitative results on LAION-Aesthetic-6+ dataset.
        With training only a few parameters on cross-attention, SUN achieves the best FID/CLIP scores above the existing adapting-free methods, the results are also competitive with SOTA methods that require finetuning the entire diffusion model.
        Guidance scale is 8.0, resolution is $512\times512$.
    }
    \small
        \begin{tabular}{l|c|c|*{3}{c}|*{3}{c}}
            \toprule
            \multirow{2}{*}{Method} & \multirow{2}{*}{Params} & \multirow{2}{*}{\makecell{Adapting \\ free}} & \multicolumn{3}{c|}{FID $\downarrow$} & \multicolumn{3}{c}{CLIP score $\uparrow$} \\
             & & & 4-step & 8-step & 12-step & 4-step & 8-step & 12-step \\
            \midrule
            DDIM \cite{song2020denoising} & 0 & \ding{51} & 22.38 & 13.83 & 12.97 & 0.258 & 0.292 & 0.315 \\
            DPM++ \cite{lu2022dpm++} & 0 & \ding{51} & 18.43 & 12.20 & 12.03 & 0.266 & 0.295 & \textbf{0.336} \\
            Guided-Distill \cite{meng2023distillation} & 860M & \ding{55} & 15.12 & 13.89 & 12.44 & 0.272 & 0.281 & 0.314  \\
            LCM \cite{luo2023latent} & 860M & \ding{55} & \textbf{11.10} & \textbf{11.84} & 12.02 & 0.286 & 0.288 & 0.320 \\
            LCM-Lora \cite{luo2023lcm} & 67.5M & \ding{51} & 16.83 & 14.30 & 13.11 & 0.271 & 0.277 & 0.319 \\
            SUN (Ours) & 18.5M & \ding{51} & 13.23 & 12.08 & \textbf{11.98} & \textbf{0.288} & \textbf{0.297} & 0.328 \\
            \bottomrule
        \end{tabular}
    \label{tab:sd15_fid} 
\end{table}

\begin{table}[tb] \centering
    \caption{
        Quantitative results on knowledge distillation FID with various pretrained diffusion models.
        Each generated samples set is compared to the corresponding ground truth set generated by 25-step-DPMSolver++ scheduler (using prompts from LAION-Aesthetic-6+).
        SUN significantly surpasses baselines in 4, 8, and 12 steps, demonstrating its ability to seamlessly switch to other diffusion models without any training.
        Guidance scale is 8.0, resolution is $512\times512$.
    }
    \small
        \begin{tabular}{l|c|c|*{2}{c}|*{2}{c}|*{2}{c}}
            \toprule
            \multirow{2}{*}{Method} & \multirow{2}{*}{Params} & \multirow{2}{*}{\makecell{Training \\ free}} & \multicolumn{2}{c|}{Rea v5.1 $\downarrow$} & \multicolumn{2}{c|}{RevA $\downarrow$} & \multicolumn{2}{c}{Any v5 $\downarrow$} \\
             & & & 4-step & 8-step & 4-step & 8-step & 4-step & 8-step \\
            \midrule
            DDIM \cite{song2020denoising} & 0 & \ding{51} & 25.32 & 21.94 & 27.22 & 22.46  & 29.88  & 23.39 \\
            DPM-Solver++ \cite{lu2022dpm++} & 0 & \ding{51} & 24.01  & 21.12  & 26.02 & 21.38  & 29.06  & 22.75 \\
            Guided-Distill \cite{meng2023distillation} & 860M & \ding{55} & 20.31 & 16.33 & 22.40 & 17.23 & 25.57 & 18.49 \\
            LCM-Lora \cite{luo2023lcm} & 67.5M & \ding{51} & 21.88 & 17.42 & 23.44 & 18.11 & 26.34 & 19.77 \\
            SUN (Ours) & 18.5M & \ding{51} & \textbf{19.60} & \textbf{15.73} & \textbf{20.27} & \textbf{16.00} & \textbf{22.52} & \textbf{16.17} \\
            \bottomrule
        \end{tabular}
    \label{tab:kd_fid}
\end{table}

\subsection{Quantitative Evaluation}

We first use SD v1.5 to test because all distillation-based acceleration methods are trained on SD v1.5.
As shown in Tab. \ref{tab:sd15_fid}, SUN contains the smallest number of parameters among all distillation methods, 
making it more efficient in training and better to reduce the risk of overfitting.
The quality of the generated images is evaluated mainly by using a standard test set as the reference. SUN is a competitive method in distribution difference (FID) and semantic consistency (CLIP score), and it achieves the best results when compared to other methods with the same parameter magnitude.

Furthermore, we evaluate the quantitative result of SUN as a universal acceleration add-on and compare it with existing techniques. 
We tested three different models that have been already fine-tuned on specialized datasets. 
As the styles of the new models are diverse, there is no standard reference set, such as LAION-5B or MSCOCO, to evaluate FIDs. 
To better reflect the consistency of the generated images before and after acceleration, for testing pretrained diffusion model, we use the 25-step DPM-Solver++ to generate 30k samples using the same prompts in Sec. \ref{sec:evaluation}, and then take them as reference for computing FID. As shown in Fig. \ref{tab:kd_fid}, SUN is demonstrated to surpass other acceleration methods on all models, therefore being a preferable acceleration method.

\begin{table}[tb] \centering
    \caption{
        Time consumption for an $512\times512$ image (seconds) using Diffusers Pipeline. \textit{Non batch parallel} puts positive prompts and negative prompts into two batches for the inference process.
    }
    \small
        \begin{tabular}{l|*{2}{c}|*{2}{c}}
            \toprule
            \multirow{2}{*}{Method (steps)} & \multicolumn{2}{c|}{V100 (FP32)} & \multicolumn{2}{c}{M1Pro (FP16)} \\
              & pipeline & unet & pipeline & unet  \\
            \midrule
            DPM-Solver++ (25) & 3.42 & 3.16 & 21.24 & 20.09 \\
            \multirow{2}{*}{\makecell{DPM-Solver++ (25) \\ (\it{non batch parallel})}} & \multirow{2}{*}{3.67} & \multirow{2}{*}{3.42} & \multirow{2}{*}{22.21} & \multirow{2}{*}{21.07} \\
             &  &  & &  \\
            DPM-Solver++ (4) & 0.684 & 0.420 & 3.97 & 2.94 \\
            \midrule
            Guided-Distill (4) & 0.459 & 0.243 & 2.42 & 1.55  \\
            \midrule
            LCM-LoRA (4) & 0.521 & 0.317 & 2.56 & 1.69 \\
            SUN (Ours) (4) & \textbf{0.485} & \textbf{0.274} & \textbf{2.50} & \textbf{1.62} \\
            \bottomrule
        \end{tabular}
    \label{tab:speed}
\end{table}

\begin{table}[t] \centering
    \caption{
      Ablative study of hyperparameter $p$ in Multi-step Consistency loss. 
      Evaluated by FID, using a excessively large value makes training difficult.
    }
    \small
        \begin{tabular}{l|cc|cc}
            \toprule
            $p$(MSC) & Rea v5.1(4) & (8) & Any v5(4) & (8) \\
            \midrule
            0.0 & 22.41 & 18.77  & 25.62  & 20.98 \\
            0.1 & \textbf{19.60} & \textbf{15.73} & \textbf{22.52} & \textbf{16.17} \\
            0.25 & 20.13  & 17.22  & 24.01  & 16.69 \\
            \bottomrule
        \end{tabular}
    \label{tab:p_fid}
  \end{table}

As an important supplement, we test the time consumption of each acceleration method on different hardware platforms (Tab. \ref{tab:speed}). SUN is faster than baseline (DPM-solver++ 25 steps) by more than 10x in terms of U-Net time consumption, and is faster than LCM-LoRA due to the advantage of parameter quantity.

\begin{figure}[tb] \centering
    \includegraphics[width =0.7\linewidth]{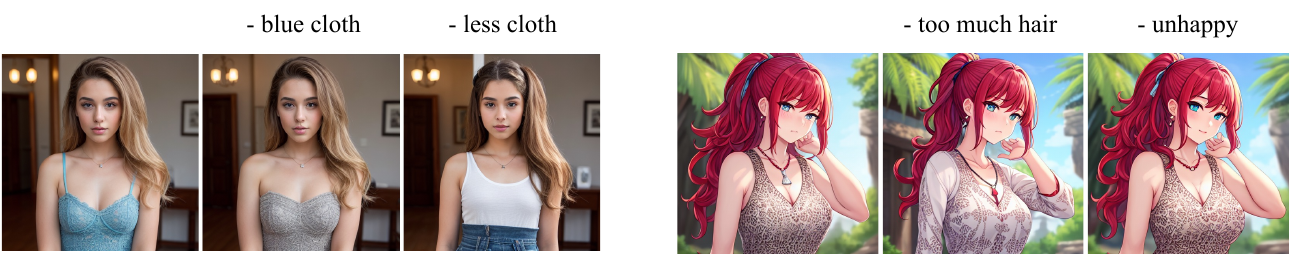}
    \caption{The proposed SUN maintains the controllability of negative prompts when eliminating the need for CFG.} 
    \label{fig:neg}
\end{figure}

\begin{figure}[tb]
  \centering
  \begin{subfigure}{0.45\linewidth}
    \includegraphics[width = \linewidth]{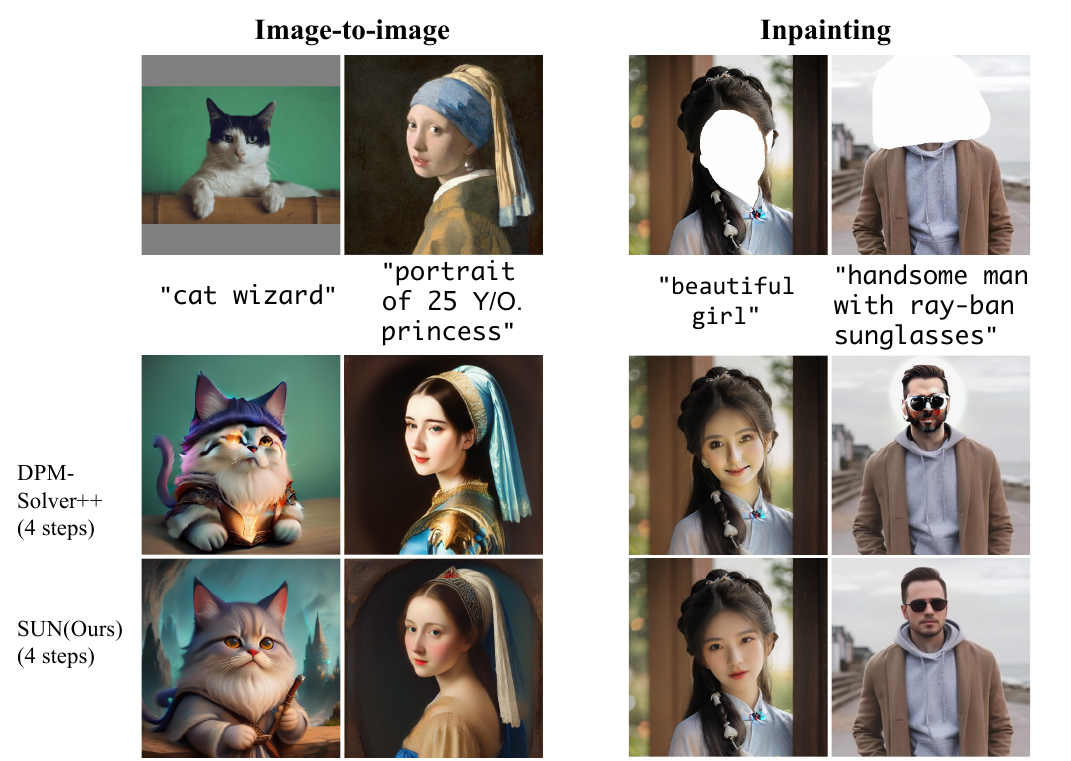}
    \caption{Image-to-image and inpainting.}
    \label{fig:img2img}
  \end{subfigure}
  \hfill
  \begin{subfigure}{0.45\linewidth}
    \includegraphics[width = \linewidth]{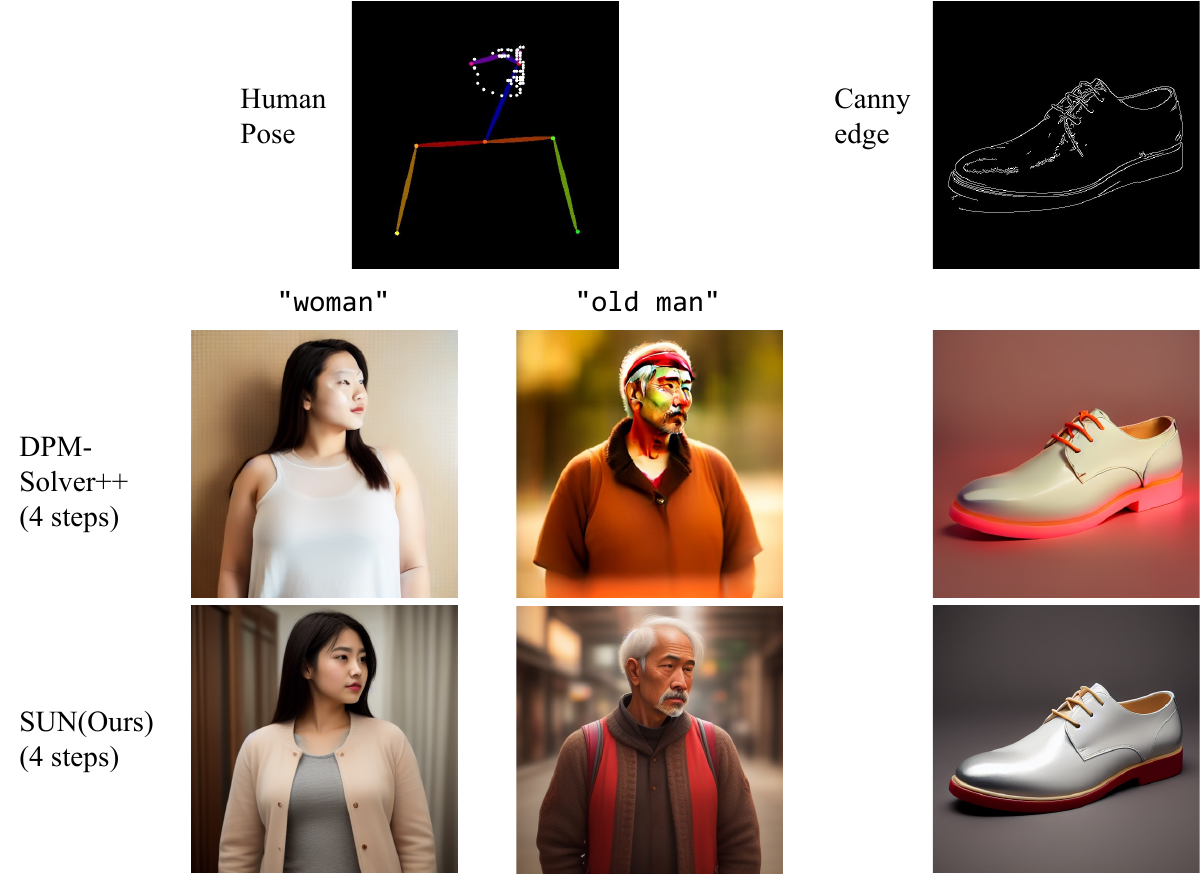}
    \caption{ControlNet.}
    \label{fig:controlnet}
  \end{subfigure}
  \caption{Without extra training, SUN can accelerate other image-generation tasks, such as inpainting and image-to-image generation. SUN is also compatible with ControlNet.}
  \label{fig:im2im_controlnet}
\end{figure}

\subsection{Other Results}

\subsubsection{Alter the Negative Prompt.}
Since the ability to modify negative prompts is essential for creators in image creation,
we further use different negative prompts for one positive prompt. The experimental results (Fig. \ref{fig:neg}) showed that SUN effectively learned the content of the negative prompt rather than fitting a specific style, achieving the same effect as CFG. 

\subsubsection{Image-to-Image and Inpainting.}
Besides text-to-image generation, SUN can also be used as a plug-in to accelerate image-to-image as well as inpaining diffusion models.
As shown in the Fig. \ref{fig:img2img}, without any training on the target model, SUN is able to generate results comparable to the original model with only 4 steps.

\subsubsection{ControlNets.}
Additional structure control is a popular application for text-to-image diffusion models.
As our SUN does not change the original network structure, it is fully compatible with
existing controllable tools (as shown in Fig. \ref{fig:controlnet}).

\begin{figure}[tb] \centering
    \includegraphics[width = 0.8\textwidth]{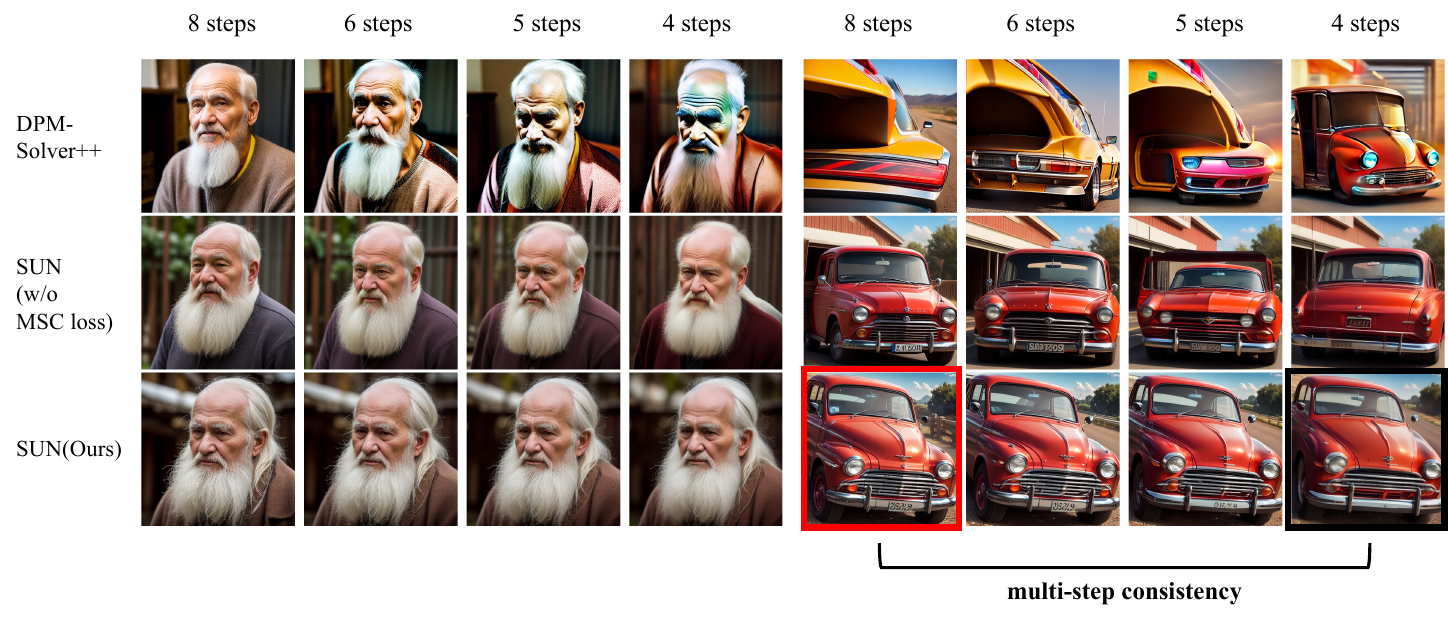}
    \caption{Ablation on our proposed Multi Step Consistency loss.
        The addition of MSC allows for the generation of samples with consistent content in 4 to 8 or more steps.
    } 
    \label{fig:ablation_mcs}
\end{figure}

\begin{figure}[t] \centering
    \includegraphics[width = 0.7\textwidth]{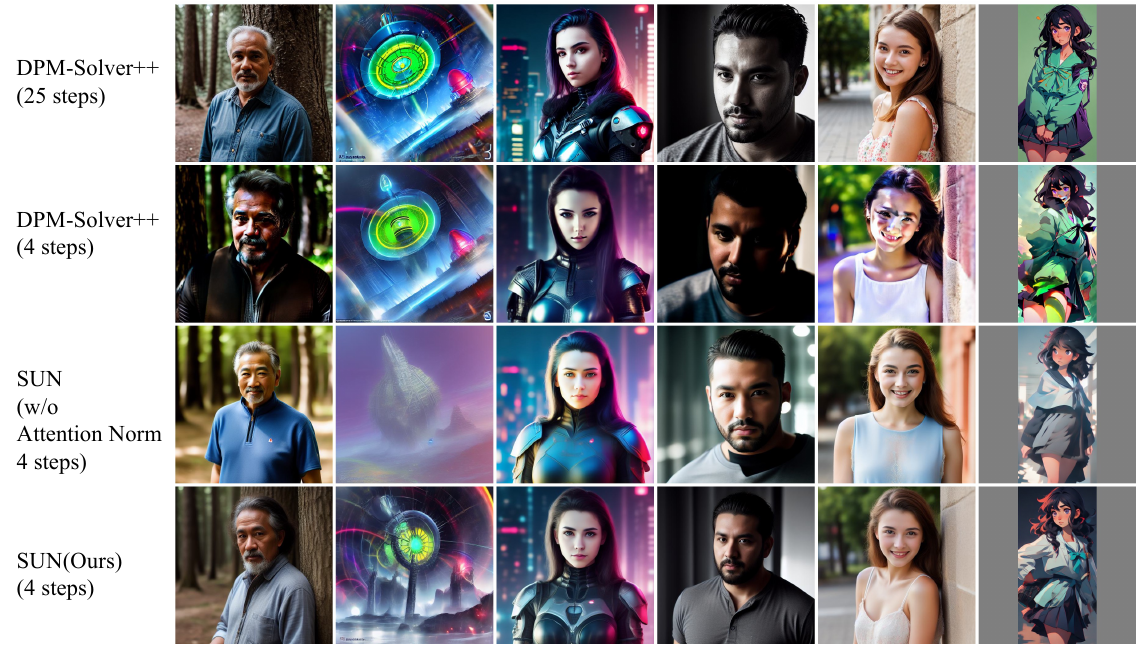}
    \caption{Ablation on our proposed Attention Normalization. 
    It enables SUN as a pluggable module to have stable generation capabilities on different pre-trained diffusion models (Realistic Vision V5.1 and Rev Animated).} 
    \label{fig:ablation_norm}
\end{figure}

\begin{figure}[tb] \centering
  \includegraphics[width = 0.65\textwidth]{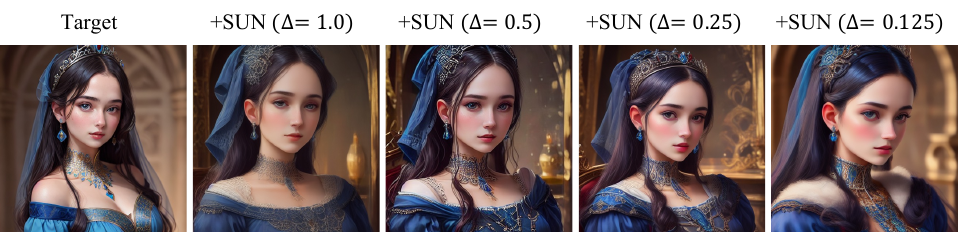}
  \caption{Ablative study of $\Delta$ in the training strategy (4 steps).
  0.25 achieves better performance in quality and consistency.}
  \label{fig:ablation_strategy}
\end{figure}

\subsection{Ablation Study}

In our research, we carried out ablation studies to evaluate two key methodological contributions of Sec \ref{sec:optimization}. Figure \ref{fig:ablation_mcs} demonstrates that MSC is crucial in ensuring that the model generates consistent content, whether in very few or multiple steps.
As shown in Fig. \ref{fig:ablation_norm}, Attention Normalization further reduces the fitting degree of SUN to the base model and helps achieve high-quality generation capabilities on different pre-trained models. 
Additionally, we do ablation studies (Fig.\ref{fig:ablation_strategy} and Tab.\ref{tab:p_fid}) to assess the impact of training strategy parameters on the results.

\section{Conclusion}
In this work, we introduced SpeedUpNet (SUN), a novel and universal Stable-Diffusion acceleration module that can be seamlessly integrated into different fine-tuned Stable-Diffusion models without further training, once it is trained on a base Stable-Diffusion model. SUN proposes a method that utilizes an adapter for the cross-attention layers in U-Net, along with a Multi-Step Consistency (MSC) loss. This approach is specifically designed to quantify and stabilize the offset in image generation caused by negative prompts relative to positive prompts.
Our empirical evaluations demonstrate that SUN significant reduces in the number of inference steps to just 4 steps and eliminates the need for classifier free guidance, which leads to a speedup of over 10 times compared to the baseline 25-step DPM-solver++, while preserving both the quality and generation consistency during the acceleration. Moreover, SUN is compatible with other generation tasks such as Inpainting \cite{meng2021sdedit} and Image-to-Image generation, enabling the use of controllable tools like ControlNet \cite{zhang2023adding}.

%
%
\bibliographystyle{splncs04}
\bibliography{main}
\end{document}